*Title*

A new method using deep learning to predict the response to cardiac resynchronization therapy

*Running Head*

Deep learning to predict CRT response

*Authors*

Kristoffer Larsen[a], BS

kalarsen@mtu.edu

Zhuo He[b], PhD

zhuoh@mtu.edu

Chen Zhao[b], MS

chezhao@mtu.edu

Xinwei Zhang[c], MD

zhangxinwei_njmu@sina.com

Quiying Sha[a], PhD

qsha@mtu.edu

Claudio T Mesquita[d], Ph.D., M.D.

claudiotinocomesquita@gmail.com

Diana Paez[e], MsEd., M.D.

d.paez@iaea.org

Ernest V. Garcia[f]. Ph.D.,

ernest.garcia@emory.edu

Jiangang Zou[c], MD, PhD, FHRS

jgzou@njmu.edu.cn

Amalia Peix[g], Ph.D., M.D.

atpeix@gmail.com

Weihua Zhou[b,h], PhD

whzhou@mtu.edu

*Institutions*

[a] Department of Mathematical Sciences, Michigan Technological University, Houghton, MI, USA

[b] College of Computing, Michigan Technological University, Houghton, MI, USA

c Department of Cardiology, The First Affiliated Hospital of Nanjing Medical University, Nanjing, Jiangsu, China

d Nuclear Medicine Department, Hospital Universitario Antonio Pedro-EBSERH-UFF, Niteroi, Brazil

e Nuclear Medicine and Diagnostic Imaging Section, Division of Human Health, Department of Nuclear Sciences and Applications, International Atomic Energy Agency, Vienna, Austria

f Department of Radiology and Imaging Sciences, Emory University, Atlanta, GA

g Nuclear Medicine Department, Institute of Cardiology, La Habana, Cuba

h Center for Biocomputing and Digital Health, Institute of Computing and Cybersystems, and Health Research Institute, Michigan Technological University, Houghton, MI, 49931, USA

*Address for correspondence*

Weihua Zhou

E-mail: whzhou@mtu.edu

Address: 1400 Townsend Drive, Houghton, MI 49931, USA

Or

Amalia Peix

E-mail: atpeix@gmail.com

Address: Nuclear Medicine Department, Institute of Cardiology, 17 No. 702, Vedado, CP 10 400, La Habana, Cuba

Or

Jiangang Zou

E-mail: jgzou@njmu.edu.cn

Address: Guangzhou Road 300, Nanjing, Jiangsu, China 210029

*Acknowledgements*

This research was supported by a Michigan Technological University Research Excellence Fund Research Seed grant (PI: Weihua Zhou), a research seed grant from Michigan Technological University Health Research Institute (PI: Weihua Zhou), and a Michigan Technological University Undergraduate Research Internship Program (PI: Kristoffer Larsen). This work was also supported by the National Natural Science Foundation of China (82070521), Clinical Competence Improvement Project of Jiangsu Province Hospital (JSPH-MA-2020-3) and Project on New Technology of Jiangsu Province (JX233C202103). The authors would like to thank the International Atomic Energy Agency (IAEA) for providing access to the data of the multicenter trial: "Value of intraventricular synchronism assessment by gated-SPECT myocardial perfusion imaging in the management of heart failure patients submitted to cardiac resynchronization therapy" (IAEA VISION-CRT), Coordinated Research Protocol E1.30.34. The authors would also like to thank the Department of Cardiology, the First Affiliated Hospital Nanjing Medical university for providing access to the data of the multicenter trial: "SPECT Guided LV Lead Placement for Incremental Benefits to CRT Efficacy" (GUIDE-CRT).

# ABSTRACT

*Background.* Clinical parameters measured from gated single-photon emission computed tomography myocardial perfusion imaging (SPECT MPI) have value in predicting cardiac resynchronization therapy (CRT) patient outcomes, but still show limitations. The purpose of this study is to combine clinical variables, features from electrocardiogram (ECG), and parameters from assessment of cardiac function with polarmaps from gated SPECT MPI through deep learning (DL) to predict CRT response.

*Methods.* 218 patients who underwent rest gated SPECT MPI were enrolled in this study. CRT response was defined as an increase in left ventricular ejection fraction (LVEF) > 5% at a 6-month follow up. A DL model was constructed by combining a pre-trained VGG16 module and a multilayer perceptron. Two modalities of data were input to the model: polarmap images from SPECT MPI and tabular data from clinical features and ECG parameters. Gradient-weighted Class Activation Mapping (Grad-CAM) was applied to the VGG16 module to provide explainability for the polarmaps. For comparison, four machine learning (ML) models were trained using only the tabular features.

*Results*. Modeling was performed on 218 patients who underwent CRT implantation with a response rate of 55.5% (n = 121). The DL model demonstrated average AUC (0.83), accuracy (0.73), sensitivity (0.76), and specificity (0.69) surpassing the ML models and guideline criteria. Guideline recommendations presented accuracy (0.53), sensitivity (0.75), and specificity (0.26).

*Conclusions*. The DL model outperformed the ML models, showcasing the additional predictive benefit of utilizing SPECT MPI polarmaps. Incorporating additional patient data directly in the form of medical imagery can improve CRT response prediction.

**ABBREVIATIONS**

CRT   Cardiac resynchronization therapy

LVEF   Left ventricular ejection fraction

MPI   Myocardial perfusion imaging

SPECT   Single-photon emission computed tomography

ML   Machine Learning

DL   Deep Learning

TL   Transfer Learning

ECG   Electrocardiogram

HF   Heart Failure

Grad-CAM   Gradient-weighted Class Activation Mapping

## INTRODUCTION

Heart failure (HF) is on the rise, between 2009 and 2012 roughly 5.7 million people had HF[1]. It is expected to increase by 46% from 2011 to 2030, resulting in a total of more than eight million people ≥18 years of age[1]. Cardiac resynchronization therapy (CRT) remains as the last resort for late-stage HF despite the significant proportion (30-40%) of patients that do not receive any benefit, i.e., CRT response[2]. The issue of CRT response stems from the selection of patients on non-optimal criteria which is complicated by the multi-faceted causes of HF. Guidelines recommend CRT for patients with low left ventricular ejection fraction (LVEF) ≤ 35%, sinus rhythm, left bundle-branch block (LBBB) with QRS duration ≥ 150 ms, and New York Heart Association (NYHA) class II, III, or ambulatory IV symptoms on guideline-directed medical therapy[3].

Machine learning (ML) allows computers to learn from data without explicit commands[4], and it can automatically discover meaningful patterns given sufficient and informative data. Deep learning (DL) exists as a subfield of ML, which utilizes neural networks, a flexible model that can directly work with large and high-dimensional data, such as text, image, and video[4]. Hence, the usage of DL allows for the direct manipulation of image data as opposed to ML. Previously, Tokodi et al.[5] has applied ML to the task of predicting all-cause mortality of patients who underwent CRT implantation. Feeny et al.[6] applied ML using 9 clinical features, such as QRS morphology, QRS duration, NYHA classification, LVEF to predict CRT response at a 6-month follow up. Apostolopoulos et al.[7] has utilized a supervised DL method to predict coronary artery disease using a VGG16 transfer learning (TL) model solely using SPECT MPI polarmaps. He et al.[8] applied the autoencoder (AE) technique, an unsupervised deep learning method, to the polarmaps of left-ventricular mechanical dyssynchrony to extract new predictors. In this study, we aimed to improve CRT response prediction using a DL method which combines both standard clinical variables and SPECT MPI polarmaps.

## METHODS

**Patient population**

This is a post hoc analysis combining data from two prospective studies: "Value of intraventricular synchronism assessment by gated-SPECT myocardial perfusion imaging in the management of HF patients submitted to cardiac resynchronization therapy" (IAEA VISION-CRT) [9] and "SPECT Guided LV Lead Placement for Incremental Benefits to CRT Efficacy" (GUIDE-CRT) [10]. VISION-CRT was a non-randomized, multinational, multicenter prospective cohort study while GUIDE-CRT was a randomized, multicenter, prospective cohort study. Patients missing with a significant number of missing features or follow up were excluded. In total, two hundred and eighteen patients with pre-CRT implantation baseline characteristics, echocardiography, and resting gated SPECT MPI were used with the CRT response defined using an increase in LVEF > 5% from the pre-implantation baseline to the 6-month follow-up. In aggregate, the patient population experienced a response rate (55.50%, n = 121) and a non-response rate (44.50%, n = 97).

**Clinical Data**

A total of 44 variables were evaluated at baseline and follow-up in this study: age, gender, and ethnicity; disease history including presence of coronary artery disease (CAD), myocardial infarction (MI), percutaneous coronary intervention (PCI), coronary artery bypass graft surgery (CABG), hypertension (HTM), diabetes (DM), and NYHA class; smoking history; and current medical treatment including patient prescribed medication. Additional variables include those derived from gated SPECT MPI.

**Gated SPECT MPI acquisition and quantification**

For VISION-CRT, gated SPECT MPI was acquired by SPECT scanners with low-energy, high-resolution collimators within seven days before CRT implantation. Resting gated SPECT scan was performed 30 minutes after injection of 740 to 1110 MBq (20-30 mCi) of 99mTc-sestamibi. Images were acquired by one-day resting gated SPECT MPI protocol with gamma cameras by 180° orbits with a complimentary 8 or 16 frames ECG-gating, according to the current guideline and a 100% (50%-150%) R-to-R gating window[11, 12]. All the images were reconstructed by the OSEM method from Emory Reconstruction Toolbox (ERTb2, Atlanta, GA) with 3 iterations, 10 subsets, power 10, and a cutoff frequency of 0.3 cycles/mm.

For GUIDE-CRT, gated SPECT MPI was performed by SPECT systems with low-energy, high-resolution collimators within seven days before CRT implantation[10]. The gated SPECT scan was performed 60-90 minutes after injection with 25-30 mCi of 99mTc-MIBI at rest. Image acquisition was performed based on a one-day resting gated SPECT MPI protocol with a dual-head or triple-head camera system, equipped with a low-energy, general-purpose collimator. The pixel size was a 64×64 matrix at least, and the zoom factor was set to 1.0. Gated images were acquired with a photopeak window of the 99mTc set as a 20% energy window centered over 140 keV. The division of the electrocardiographic R-R interval was 8 frames per cardiac cycle, using a 50% beat acceptance window. All the images were reconstructed using the system-equipped iterative reconstruction algorithms.

All the short-axis and planar projection images in both VISION-CRT and GUIDE-CRT were quantified by Emory Cardiac Toolbox (ECTb4, Atlanta, GA) for automated measurement of LV function and LV mechanical dyssynchrony (LVMD) including left ventricular ejection fraction (LVEF), left ventricular end-systolic volume (ESV), myocardial mass, stroke volume, wall thickening (WT), summed rest score (SRS), and scar; phase systolic and diastolic parameters including phase peak (PP), phase standard deviation (PSD), phase kurtosis (PK), phase skew (PS), and phase bandwidth (PBW); and shape parameters including end systolic eccentricity (ESE), end diastolic eccentricity (EDE), and shape index (ESSI and EDSI). The concordance, defined as the agreement between CRT LV lead position used and the optimal LV lead position identified by ECTb4 as the latest contracting viable site, was considered[13, 14].

Three rest polarmap images from gated SPECT MPI were extracted from *Emory Cardiac Toolbox* including perfusion, systolic dyssynchrony, and systolic wall thickening polarmaps. A fourth polarmap image was generated by modifying the perfusion polarmap by thresholding to only include pixel values greater than a certain value on a scale of [0,1]. In this study a threshold of 0.50 was used to include pixels as a middle point on the pixel value scale. As seen in Figure 1, the polarmap images were concatenated together to form a 128 by 128-pixel image. The overall image is oriented in order by the perfusion ("upper-left"), systolic dyssynchrony ("upper-right"), wall thickening ("bottom-left"), and then the modified perfusion polarmap ("bottom-right").

**DL and ML models**

Our DL model is a supervised multi-input model which implements TL through a VGG16 model to extract information from the polarmap images. TL is a technique used to overcome the issue of limited training data. In convention, TL is used to improve the understanding of a specific task by relating it to other tasks which have much more abundant data[15]. In this study, the VGG16 model trained on the ImageNet dataset[16] is utilized.

Through TL by the VGG16 module, the overall DP multi-input model can extract features from polarmap images. For the tabular features, a multi-layer perceptron (MLP) is utilized. MLP is the most widely used neural network structure which contains multiple hidden layers that non-linearly transform inputs so that the output of the model is optimized towards a certain task[17]. The overall DL model architecture for this study is shown in figure 2. The model parameters were updated via the Adam optimizer[18]. The overall model was implemented through the TensorFlow package in the Python programming language (Version 2.7.0) while hyperparameter tuning of the model was performed via Keras-Tuner (Version 1.1.3). The model was trained on graphical processor units (NVIDIA Tesla V100).

Feature selection was performed to select relevant patient variables to be used in the models. Recursive Feature Elimination (RFE) was performed with 7-fold cross validation to automatically select the optimal feature subsets. The features used in each fold are shown in Table 1. Afterwards, the respective subset of features was passed through a correlation filter using Pearson correlation (≤ 0.80) to reduce redundancy and a near-zero variance filter (threshold 0.01) to remove near constant variables. Finally, the features were centered and scaled to be optimal inputs for the neural network.

A nested cross validation structure was employed to validate the models' performance. A 5-fold stratified shuffle split with (10/11, n = 198) of the samples for training split and (1/11, n = 20) of the samples for the test splits was employed for the outer fold. For the inner fold to hyperparameter tune the model, a 4-fold stratified shuffle split with (9/10, n = 178) of the samples for the training split and (1/10, n = 20) of the samples for the test split. Bayesian hyperparameter tuning was utilized to select the optimal hyperparameter configuration of the models. For the image data, the grayscale polarmap images were scaled to the range [0,255] and tiled three times along the channel dimension before centering RGB-wise according to the ImageNet dataset as specified by the VGG16 module.

For performance comparison, four ML models were considered solely using the tabular data, specifically, the clinical features. The four models include Logistic Regression (LR), Random Forest (RF), Adaboost Decision Tree (ADA), and Support Vector Machine (SVM). The ML models were trained using the same nested-cross validation structure, features, and data preprocessing steps as the DL model.

Additionally, based on the guideline recommendations of the 2012 ACCF/AHA/HRS, another classifier was constructed[19]. Patients with Class I or Class IIa CRT recommendations were predicted as CRT responders. Performance metrics including accuracy, sensitivity, and specificity were calculated from a confusion matrix using this guideline method.

Grad-CAM is a technique for providing explainability to CNN models by producing a coarse localization map highlighting important regions in the image for predicting the target class[20]. Heat maps generated by Grad-CAM can be overlaid onto the original input image to provide explainability for the model's choice of prediction. In this study, we perform Grad-CAM to gain insight into how the DL classifier places importance amongst the different SPECT MPI polarmaps and their respective patterns exhibited by different patients.

**Statistical analysis**

Categorical variables at baseline were expressed in counts and percentages with p-values provided through the chi-square test of independence. Continuous variables were expressed as mean $\pm$ standard deviation with p-values provided through the two-sample t-test. The predictive performance of the different models was evaluated using Area Under the Curve (AUC), accuracy, sensitivity, and specificity.

## RESULTS

A total of 218 patients underwent CRT implantation, completing both a baseline SPECT MPI and clinical assessment and a 6-month follow-up. The baseline patient data is shown in Table 2. After a 6-month follow-up, a total of 121 out of 218 patients (55.5%) were classified as responders defined as an increase in LVEF > 5%.

Average performance metrics across the outer validation folds are presented in Table 3 for each of the models. For CRT response prediction, the DL multi-input model outperformed all other models in comparison. The DL model presented an AUC of 0.83, accuracy of 0.73, sensitivity of 0.76, and specificity of 0.69. In comparison to the ML models, the ENET model presented the next highest performance with an AUC of 0.82, accuracy of 0.71, sensitivity of 0.75, and specificity of 0.67. Next in terms of performance was the SVM model with an AUC of 0.80, accuracy of 0.66, sensitivity of 0.69, and specificity of 0.62. Below the SVM in terms of performance is the ADA model with AUC of 0.75, accuracy of 0.65, sensitivity of 0.69, and specificity of 0.60. The lowest performing ML model was RF with an AUC of 0.74, accuracy of 0.61, sensitivity of 0.64, and specificity of 0.58. The DL model had a sizable increase in performance against the guideline model with accuracy 0.73 vs. 0.53, sensitivity 0.76 vs. 0.75, and specificity 0.69 vs. 0.26, respectively.

For the standard deviation among the folds, the RF model exhibits the lowest value for AUC 0.06 against the next lowest of DL 0.07. For accuracy, both the ENT and RF model were tied for the lowest standard deviation at 0.06, respectively. The DL model is ranked third below the SVM model at 0.09 vs. 0.07, respectively. For sensitivity, the RF model again displayed the lowest standard deviation of 0.06, the next lowest was that of SVM at 0.09, and thirdly the DL model with 0.11. The DL model exhibited the lowest standard deviation for specificity.

**Explainability to the DL model**

For explainability, Grad-CAM was performed to extract the DL model's regional choice of importance in the SPECT MPI polarmaps with respect to the classification task of CRT response.

From the nested validation, the best performing model in the outer fold in terms of AUC (Fold 1) was selected. Grad-CAM analysis is presented for 4 patients in the testing fold, including the original polarmap inputs to the model and the overlaid polarmap heatmap in Figure 3.

Patient (A) in Figure 3 (true positive) is a class I guideline recommendation patient expressing regions of interest in the perfusion, systolic dyssynchrony, and wall thickening polarmap. Though minor, there is some heat exhibited by Grad-CAM across the apex region of the perfusion polarmap. In contrast, there is a significant coloring display on the lateral side of the systolic dyssynchrony polarmap over the heterogenous contractile pattern. On the wall-thickening polarmap, interest is shown in the anterior region where there is an abnormal thickening pattern. Ultimately, the patient is correctly predicted as a responder.

Patient (B) in Figure 3 (true negative) is a class IIa guideline recommendation displaying significant regions of interest in the systolic dyssynchrony ("upper-right') and wall thickening ("bottom-left") polarmaps. The entire systolic dyssynchrony polarmap which possesses a heterogeneous contractile pattern is identified; while the wall thickening polarmap with an abnormal thickening pattern in the basal inferior and mid-inferior. Ultimately, the patient is correctly predicted as being a non-responder.

Patient (C) in Figure 3 (false negative) is a class I guideline recommendation exhibiting a slight perfusion defect in the apex region as seen in the modified perfusion polarmap ("bottom-left"). The overlaid heatmap identifies this region shown by the strong yellow and red colors, however, little attention is paid to the other polarmaps in this patient. Ultimately, the patient is incorrectly predicted as being a responder.

Patient (D) in Figure 3 (false positive) is a class I guideline recommendation with regions of interest in the scatter in the anterior region of the perfusion polarmap. Moreover, strong regions of interests are placed in the wall thickening polarmap along the anteroseptal regions. Correspondingly where perfusion polarmap is colored, in the modified perfusion polarmap where this region is excluded there is no interest colored. Ultimately, the patient is incorrectly predicted as being a non-responder.

## DISCUSSION

In this study, we developed a TL multi-input model to predict CRT patient response. The DL model using clinical parameters and SPECT MPI polarmaps, including myocardial variability, systolic dyssynchrony, and systolic wall thickening polarmaps, were validated and compared to conventional ML models using only tabular data from VISION-CRT and GUIDE-CRT trials. The

DL model outperformed both the ML models and current guideline recommendations for admission. In comparison of the DL model against the highest performing ML model (ENET), the DL model demonstrated an AUC of 0.83 vs. 0.82 and accuracy of 0.73 vs. 0.71, respectively, showing the improved predictive ability of incorporating SPECT MPI polarmaps. Although not statistically significant, the DL model trended towards improved performance. Moreover, compared with guideline criteria for recommendation, the DL showed massive improvement with accuracy 0.73 vs. 0.53, sensitivity 0.76 vs. 0.75, and specificity 0.69 vs. 0.26, respectively. Although modest, incorporating SPECT MPI polarmaps can improve prediction performance for CRT response. CRT response is complex to predict due to the multifaceted nature of HF[21], but through incorporating additional informative patient data directly with derived parameters through feature engineering clinical decision-making support predictions can be improved. Incorporating raw data directly through DL represents an enormous development for the treatment of HF and the associated clinical decision making process[22, 23]. Moreover, while the sensitivity of the guideline method is similar to that of the DL method, the low specificity of the guideline approach should be noted (0.26). This low specificity originates from an excess of inserted CRTs that did not improve patient outcomes. Furthermore, the low specificity of the guideline approach (0.26) highlights the need to better select patients that display higher suitability for improved outcomes after CRT and to avoid inserting CRTs when there is a lower indication of a positive CRT response. This approach can reduce the associated patient cost and wasted time for patients experiencing HF.

Tabular features in the form of standard clinical data, cardiac functional parameters, shape parameters, LVMD data from gated SPECT MPI, and QRSd from ECG are informative for multivariate analysis for the prediction of CRT response. In addition, the inclusion of SPECT MPI polarmap was found to enhance the predictive ability when fused with tabular features.

The DL model which incorporates SPECT polarmaps images provides explainability in the model's predictions though Grad-CAM heatmaps. DL models are often considered "black boxes" due to the unexplainable complexity which is not easily understand by humans[24]; despite this concern, many techniques exist to improve the clinical decision-support of these black box models, such as Grad-CAM. Existing as a local explainable technique, Grad-CAM allows clinicians the opportunity to delve into why the DL model forms its prediction on a patient-to-patient level. Moreover, since the DL model fuses both the polarmap images and the clinical features, the explainability in the form of Grad-CAM heatmaps omits information regarding how the tabular features blend with the polarmap images to affect the resulting CRT response prediction.

One of the strengths of this study was the usage of patient data from the VISION-CRT and GUIDE-CRT trials which provided multicenter and multinational data. For VISION-CRT, 10 centers across 8 different countries provided patient data including Brazil, Chile, Columbia, Cuba, India, Mexico, Pakistan, and Spain. For GUIDE-CRT, 19 centers across China provided patient data. As such, combining the data from these two studies supplies a wider range of many different population groups, not only increasing the representativeness, but also the generalizability of this analysis.

**Limitations and Future Work**

This study performed analysis on a relatively small number of patients (218). In addition, this study performed analysis from centers which all used the *Emory Cardiac Toolbox* to generate both the SPECT MPI polarmaps and derived variables. Differences in software implementations to calculate LV function and LVMD have the potential to produce different results[29, 30, 31]. Hence, special care is required when comparing results from different available commercial software packages for MPI quantification.

### New Knowledge Gained

In this study, we employed the use of TL to circumvent the issues of limited samples from multi-center data. Additionally, Grad-CAM was applied to the DL model to explain our findings on the SPECT MPI polarmaps. DL showed improvement over ML, while both DL and ML demonstrated improvement over guideline recommendations.

### Conclusions

This paper proposed a method using DL to predict CRT response combining clinical parameters and polarmap images from gated SPECT MPI. This multi-input DL model outperforms the use of ML solely using clinical parameters and current guideline recommendations for predicting CRT response.

### Conflict of Interest Disclosure Statement

All authors declare that there are no conflicts of interest.

Figure 1. Polarmap input for DL model. Polarmaps were generated from pre-CRT gated SPECT MPI, including myocardial perfusion ("upper-left"), systolic dyssynchrony ("upper-right"), wall thickening ("bottom-left"), and then the modified polarmap with perfusion defect ("bottom-right").

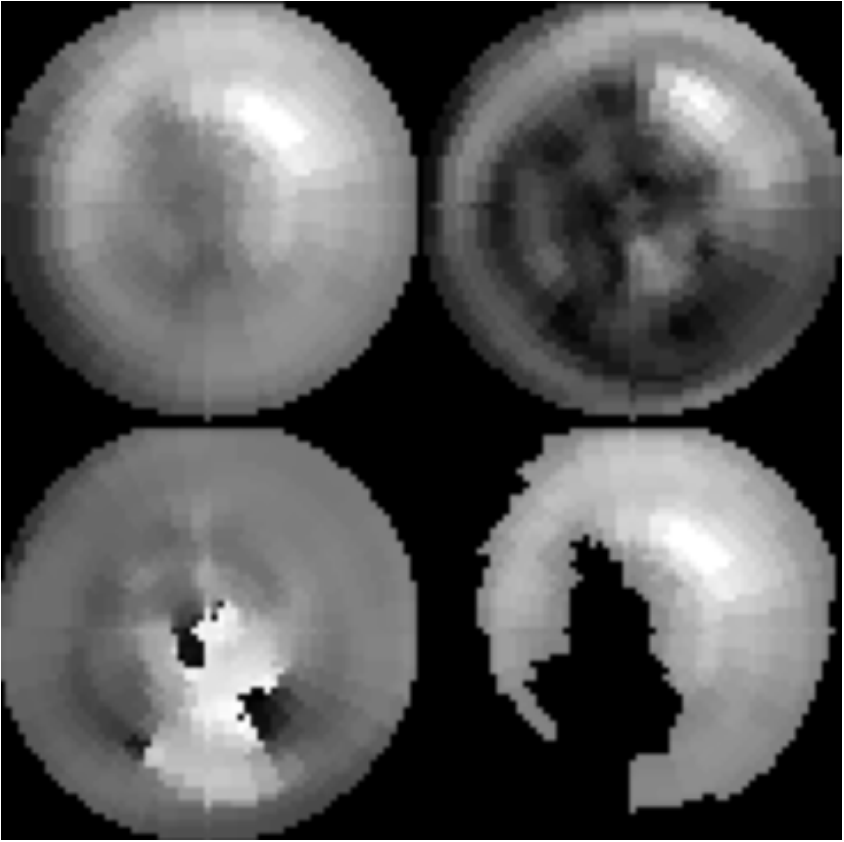

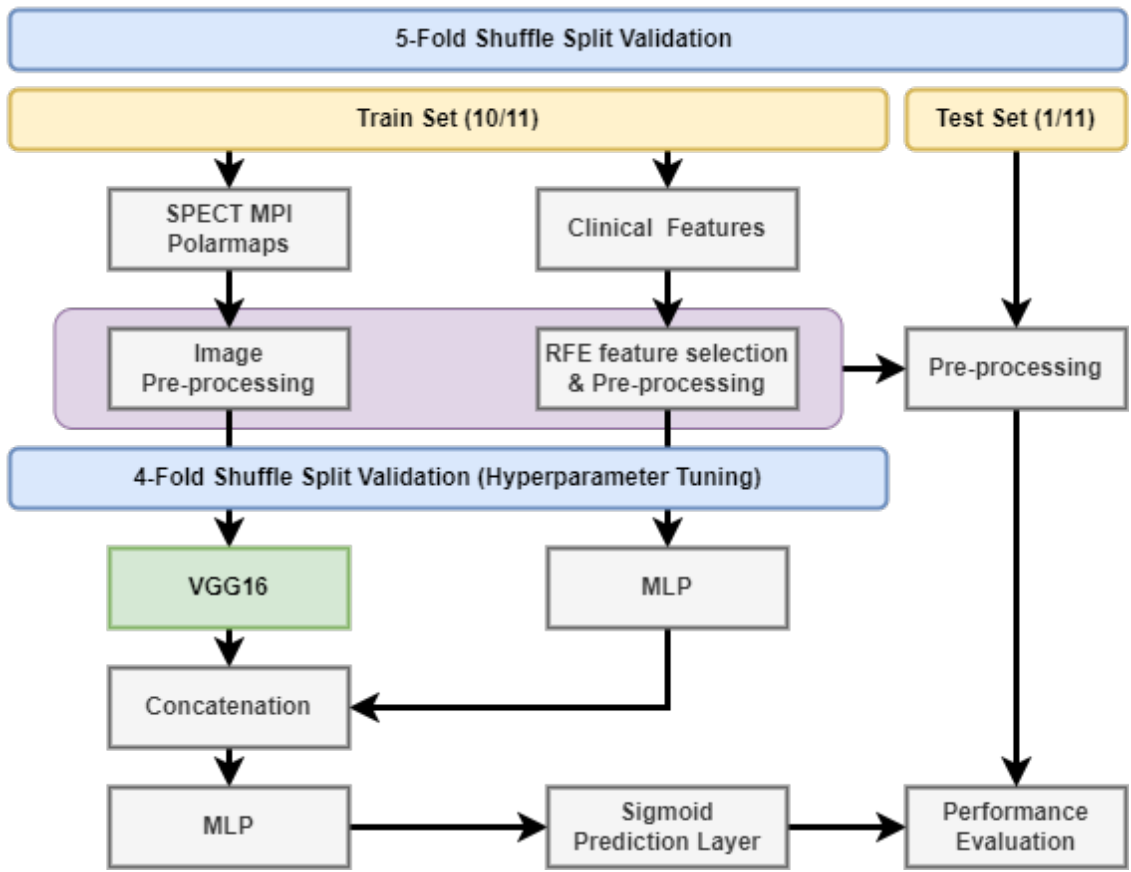

Figure 2. DL model overview.

Figure 3. Grad-CAM images. (A) Class I patient, True Positive. (B) Class IIa patient, True Negative. (C) Class I patient, False Positive. (D) Class I patient, False Negative.

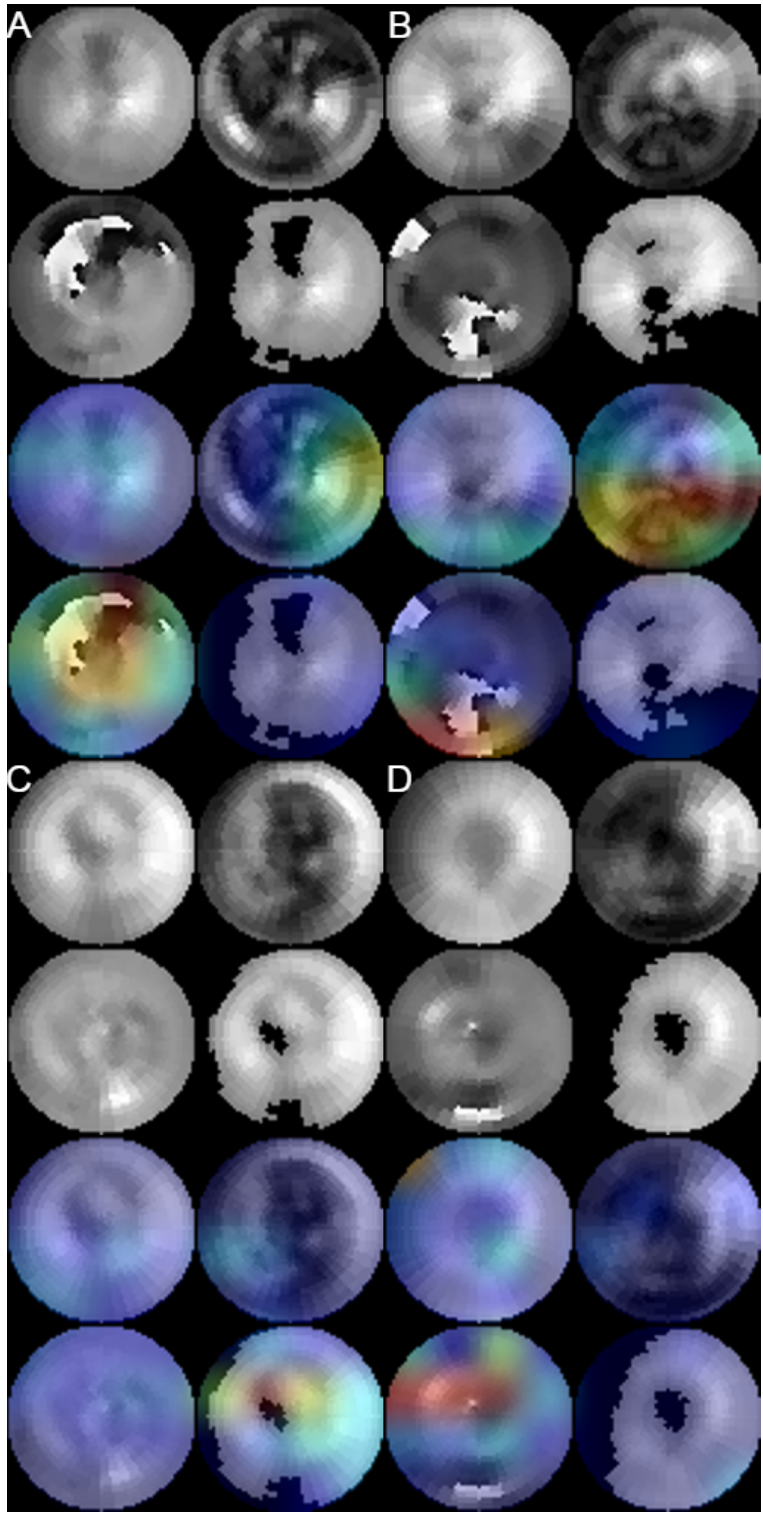

Figure 4. AUC curves. In reading order: DL, Support Vector Machine (SVM), Adaboost Decision Tree (ADA), Random Forest (RF), and Elastic Net Logistic Regression (ENET)

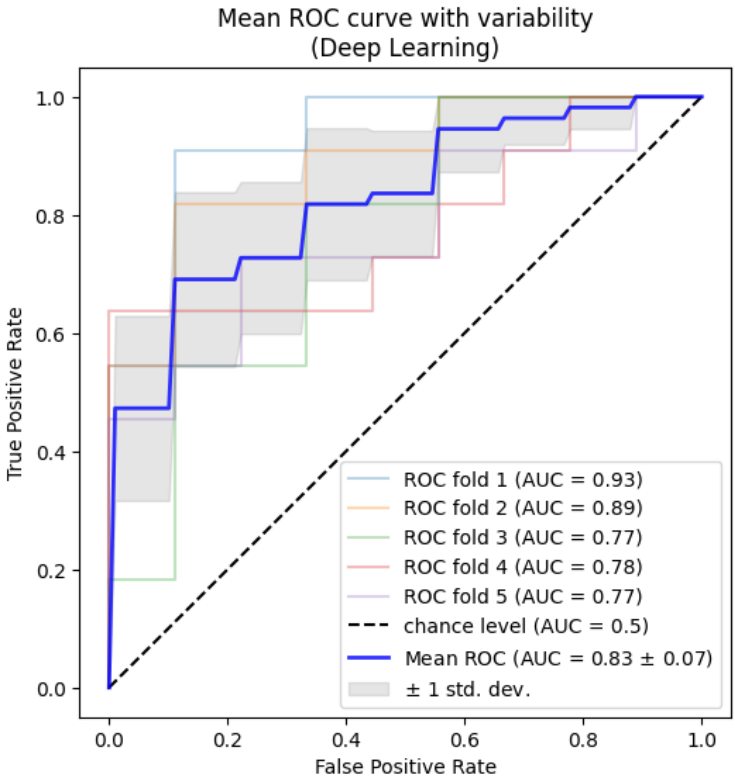
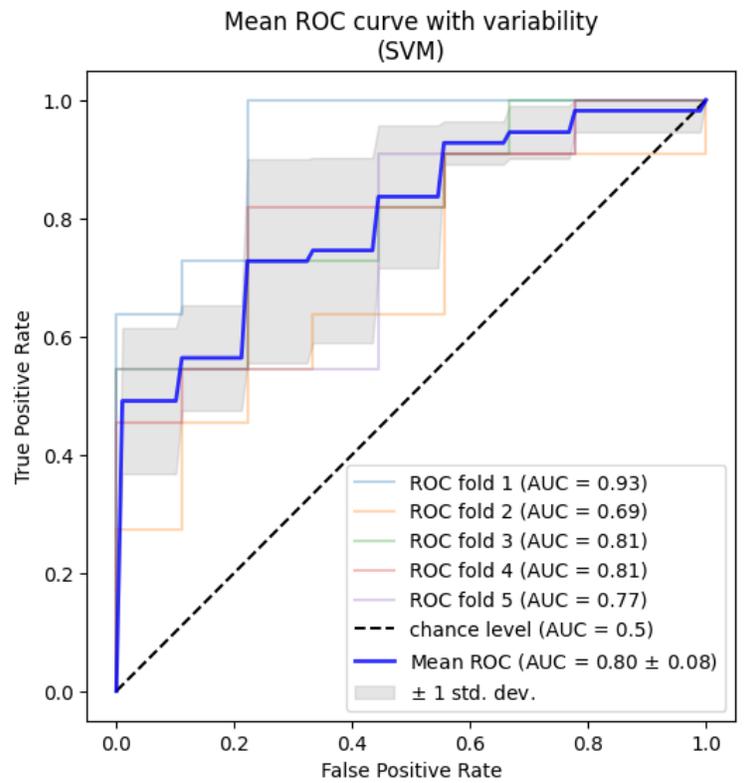
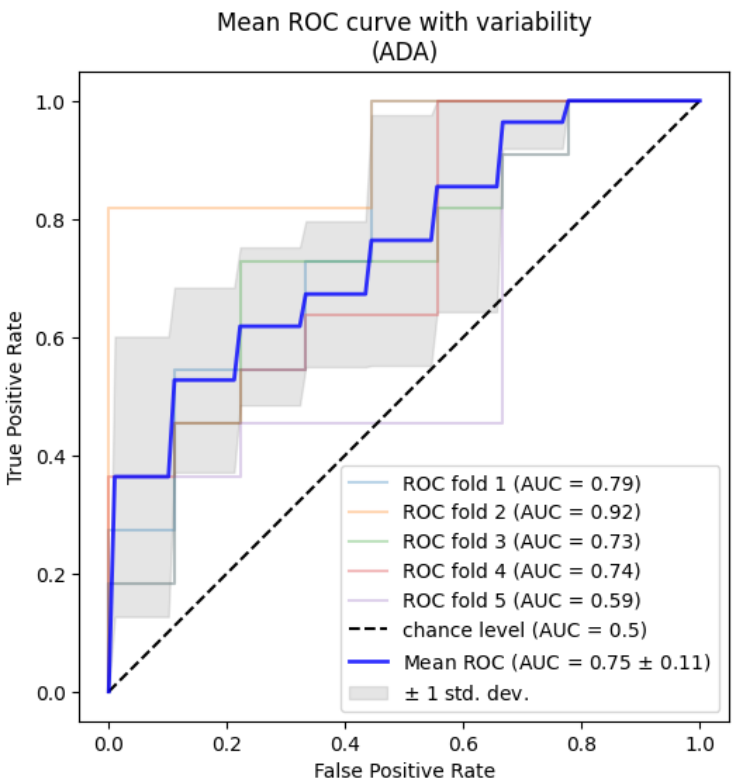
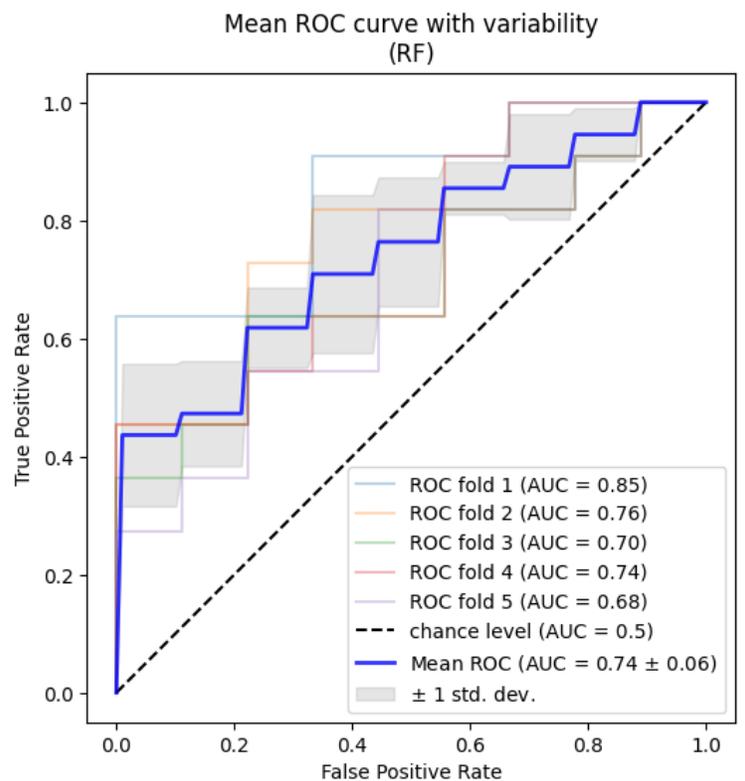

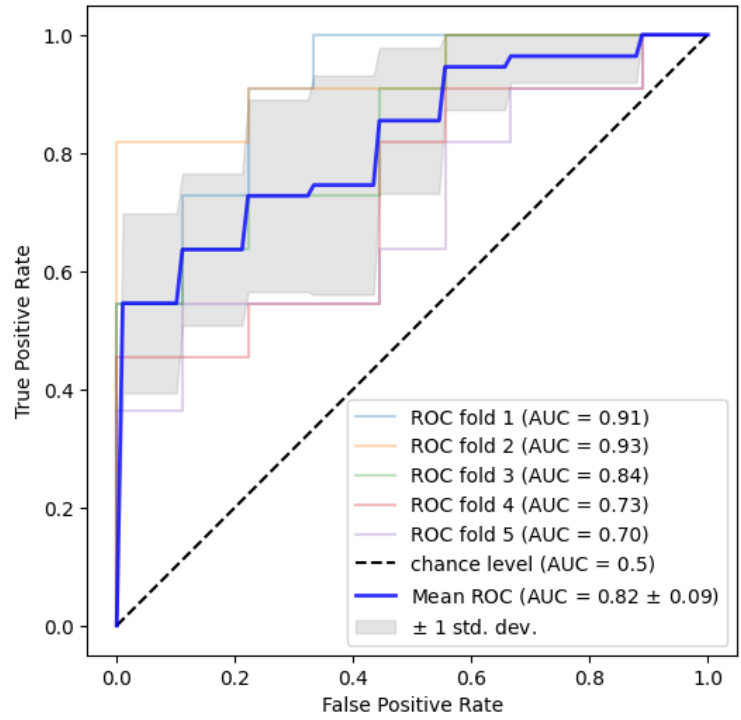

Table 1. Features used for modeling.

| Fold | Clinical | SPECT/ECG |
|---|---|---|
| **1** (N = 11) | CABG<br>**CAD**<br>NYHA 2<br>**Race Asian**<br>**Race Hispanic**<br>**Smoking** | Scar %<br>**EDV**<br>**ESE**<br>**ESSI**<br>**LVEF** |
| **2** (N = 33) | ACEI/ARB<br>Age<br>CABG<br>**CAD**<br>DM<br>Gender<br>HTN<br>LBBB<br>MI<br>NYHA 2<br>NYHA 3<br>NYHA 4<br>PCI<br>Race African<br>**Race Asian**<br>Race Caucasian<br>**Race Hispanic**<br>Race Indian<br>**Smoking** | Concordance<br>QRSd<br>Scar %<br>Diastolic PK<br>**EDV**<br>**ESE**<br>**ESSI**<br>**LVEF**<br>Systolic PBW<br>Systolic PK<br>Systolic PP<br>SRS<br>Stroke Volume<br>WT Sum |
| **3** (N = 26) | ACEI/ARB<br>**CAD**<br>DM<br>Gender<br>MI<br>NYHA 2<br>NYHA 3<br>NYHA 4<br>PCI<br>Race African<br>**Race Asian**<br>Race Caucasian<br>**Race Hispanic**<br>Race Indian<br>**Smoking** | Diastolic PBW<br>Diastolic PS<br>**EDV**<br>**ESE**<br>**ESSI**<br>**LVEF**<br>Systolic PP<br>SRS<br>Stroke Volume |
| **4** (N = 28) | ACEI/ARB<br>Age<br>CABG<br>**CAD**<br>DM<br>Gender<br>LBBB<br>NYHA 3<br>NYHA 4<br>PCI<br>Race African<br>**Race Asian**<br>Race Caucasian<br>**Race Hispanic**<br>Race Indian<br>**Smoking** | QRSd<br>Scar %<br>Diastolic PK<br>**EDV**<br>**ESE**<br>**ESSI**<br>**LVEF**<br>Systolic PBW<br>Systolic PP<br>SRS<br>Stroke Volume<br>WT Sum |
| **5** (N = 31) | ACEI/ARB<br>Agee<br>**CAD**<br>DM<br>Gender<br>HTN<br>LBBB<br>MI<br>NYHA 2<br>NYHA 3<br>NYHA 4<br>PCI<br>Race African<br>**Race Asian**<br>Race Caucasian<br>**Race Hispanic**<br>Race Indian<br>**Smoking** | QRSd<br>Diastolic PK<br>**EDV**<br>**ESE**<br>**ESSI**<br>**LVEF**<br>Systolic PBW<br>Systolic PK<br>Systolic PP<br>SRS<br>Stroke Volume<br>WT Sum |

In **bold** are features appearing in every fold.

DM, Diabetes Mellitus; HTN, Hypertension; MI, Myocardial Infarction; CAD, Coronary Artery Disease; CABG, Coronary Artery Bypass Graft surgery; PCI, Percutaneous Coronary Intervention; NYHA, New York Heart Association; ACEI/ARB, Angiotensin-Converting-Enzyme Inhibitors/Angiotensin Receptor Blockers; SRS, Summed Rest Score; ESV, End-Systolic Volume; WT, wall thickening; Scar %, percentage non-viable LV; PBW, Phase Bandwidth; PK, Phase Kurtosis; PS, Phase Skew; PP, Phase Peak; PSD, Phase Standard Deviation; EDE, End-Diastolic Eccentricity; EDSI, End-Diastolic Shape Index; EDV, End-Diastolic Volume; ESE, End-Systolic Eccentricity; ESSI, End-Systolic Shape Index.

Table 2. Baseline characteristics of the enrolled patients.

| Variable | Overall (n=218) | Response (n=89, 68.5%) | Non-response (n=41, 31.5%) | P-value |
|---|---|---|---|---|
| **Age (yrs)** | 62.0 ± 11.8 | 60.5 ± 12.1 | 63.2 ± 11.5 | 0.098 |
| **Male** | 122 (61.0%) | 63 (64.9%) | 70 (57.9%) | 0.353 |
| **Race** | | | | |
|    African | 15 (6.9%) | 9 (9.3%) | 6 (5.0%) | 0.326 |
|    Asian | 76 (34.9) | 21 (21.6%) | 55 (45.5%) | <.0001* |
|    Caucasian | 21 (9.6%) | 11 (11.3%) | 10 (8.3%) | 0.593 |
|    Hispanic | 76 (34.9%) | 45 (46.4%) | 31 (25.6%) | 0.002* |
|    Indian | 30 (13.7%) | 11 (11.3%) | 19 (15.7%) | 0.465 |
| **Smoking** | 41 (18.8%) | 18 (18.6%) | 23 (19.0%) | 1.0 |
| **DM** | 53 (24.3%) | 27 (27.8%) | 26 (21.5%) | 0.354 |
| **HTN** | 117 (53.7%) | 57 (58.8%) | 60 (49.6%) | 0.225 |
| **MI** | 34 (15.6%) | 23 (23.7%) | 11 (9.1%) | 0.006* |
| **CAD** | 68 (31.2%) | 41 (42.3%) | 27 (22.3%) | 0.003* |
| **CABG** | 3 (1.4%) | 1 (1.0%) | 2 (1.7%) | 1.0 |
| **PCI** | 12 (5.5%) | 7 (7.2%) | 5 (4.1%) | 0.488 |
| **NYHA** | | | | |
|    II | 59 (27.1%) | 21 (21.6%) | 38 (31.4%) | 0.145 |
|    III | 127 (58.3%) | 65 (67.0%) | 62 (51.2%) | 0.027* |
|    IV | 32 (14.7%) | 11 (11.3%) | 21 (17.4%) | 0.292 |
| **ACEI/ARB** | 179 (82.1%) | 73 (75.3%) | 106 (87.6%) | 0.029* |
| **ECG QRSd** | 158.6 ± 27.2 | 157.8 ± 27.3 | 159.2 ± 27.1 | 0.718 |
| **SPECT** | | | | |
|    SRS | 18.2 ± 12.2 | 21.6 ± 12.5 | 15.5 ± 11.3 | <.0001* |
|    ESV | 192.6 ± 108.0 | 207.1 ± 118.7 | 181.0 ± 97.5 | 0.082 |
|    LVEF | 27.7 ± 11.0 | 28.8 ± 11.1 | 26.9 ± 10.9 | 0.207 |
|    Mass | 215.2 ± 58.4 | 224.3 ± 61.6 | 207.9 ± 54.9 | 0.042* |
|    Stroke | 62.9 ± 23.2 | 71.0 ± 24.1 | 56.4 ± 20.1 | <.0001* |
| **Volume** | | | | |
|    WT % | 22.0 ± 16.9 | 24.9 ± 16.6 | 19.8 ± 16.9 | 0.026* |
|    WT Sum | 11.2 ± 8.7 | 12.7 ± 8.5 | 10.1 ± 8.6 | 0.027* |
|    Concordance | 49 (22.5%) | 25 (25.8%) | 24 (19.8%) | 0.379 |
|    Scar % | 22.7 ± 14.0 | 27.0 ± 16.0 | 19.2 ± 11.1 | <.0001* |
|    Diastolic | | | | |
|      PBW | 169.4 ± 80.5 | 171.6 ± 78.9 | 167.7 ± 82.0 | 0.725 |
|      PK | 8.4 ± 7.3 | 8.8 ± 7.8 | 8.1 ± 6.8 | 0.456 |
|      PS | 2.5 ± 0.8 | 2.5 ± 0.9 | 2.5 ± 0.8 | 0.819 |
|      PP | 221.0 ± 0.8 | 220.6 ± 34.2 | 221.4 ± 43.7 | 0.886 |
|      PSD | 52.3 ± 20.7 | 52.7 ± 20.0 | 52.0 ± 21.4 | 0.803 |
|    Systolic | | | | |
|      PBW | 158.3 ± 77.5 | 159.8 ± 77.3 | 157.1 ± 77.9 | 0.798 |
|      PK | 132.5 ± 8.1 | 9.2 ± 9.3 | 7.5 ± 6.9 | 0.134 |
|      PP | 132.5 ± 37.2 | 131.0 ± 32.8 | 133.6 ± 40.4 | 0.599 |
|      PSD | 50.0 ± 20.6 | 50.5 ± 20.8 | 49.7 ± 20.5 | 0.758 |
|    EDE | 0.6 ± 0.2 | 0.5 ± 0.2 | 0.6 ± 0.1 | 0.002* |
|    EDSI | 0.8 ± 0.1 | 0.8 ± 0.1 | 0.7 ± 0.1 | 0.003* |
|    EDV | 255.5 ± 118.0 | 278.2 ± 129.4 | 237.4 ± 105.1 | 0.013* |
|    ESE | 0.6 ± 0.2 | 0.6 ± 0.2 | 0.6 ± 0.1 | <.0001* |
| **ESSI** | 0.8 ± 0.1 | 0.8 ± 0.1 | 0.8 ± 0.1 | 0.001* |

Data are expressed as mean ± standard deviation or count (percentage)

* P-value < 0.05

DM, Diabetes Mellitus; HTN, Hypertension; MI, Myocardial Infarction; CAD, Coronary Artery Disease; CABG, Coronary Artery Bypass Graft surgery; PCI, Percutaneous Coronary Intervention; NYHA, New York Heart Association; ACEI/ARB, Angiotensin-Converting-Enzyme Inhibitors/Angiotensin Receptor Blockers; SRS, Summed Rest Score; ESV, End-Systolic Volume; WT, wall thickening; Scar %, percentage non-viable LV; PBW, Phase Bandwidth; PK, Phase Kurtosis; PS, Phase Skew; PP, Phase Peak; PSD, Phase Standard Deviation; EDE, End-Diastolic Eccentricity; EDSI, End-Diastolic Shape Index; EDV, End-Diastolic Volume; ESE, End-Systolic Eccentricity; ESSI, End-Systolic Shape Index.

Table 3. Performance of DL and ML models.

| Model | AUC | Accuracy | Sensitivity | Specificity |
|---|---|---|---|---|
| multi-input DL | **0.83** (0.07) | **0.73** (0.09) | **0.76** (0.11) | **0.69** (**0.11**) |
| ENET | 0.82 (0.09) | 0.71 (**0.06**) | 0.75 (0.12) | 0.67 (0.14) |
| SVM | 0.80 (0.08) | 0.66 (0.07) | 0.69 (0.09) | 0.62 (0.15) |
| ADA | 0.75 (0.11) | 0.65 (0.14) | 0.69 (0.14) | 0.60 (0.21) |
| RF | 0.74 (**0.06**) | 0.61 (**0.06**) | 0.64 (**0.06**) | 0.58 (0.13) |
| Guideline | N/A | 0.53 | 0.75 | 0.26 |

Performance metrics are presented as mean (standard deviation). **Bold** represent best performance values.

ENET = Elastic Net Logistic Regression, SVM = Support Vector Machine, ADA = Adaboost Decision Tree, RF = Random Forest